\documentclass[letterpaper]{article} 
\usepackage{aaai2027}  
\usepackage[hyphens]{url}  
\usepackage{graphicx} 
\usepackage{natbib}  
\usepackage{caption} 
\usepackage{algorithm}
\usepackage{algorithmic}
\usepackage{array}
\usepackage{newfloat}
\usepackage{listings}
\DeclareCaptionStyle{ruled}{labelfont=normalfont,labelsep=colon,strut=off} 
\floatstyle{ruled}
\newfloat{listing}{tb}{lst}{}
\floatname{listing}{Listing}

\usepackage{booktabs}
\usepackage[table]{xcolor}

\usepackage{amsmath,amssymb}

\newcommand{\method}{LogFloor}

\title{Small Models Scout Bottleneck Order for Large-Model Data Control}
\author{
Seungmin Choi\textsuperscript{\rm 1},
Jiwon Sung\textsuperscript{\rm 1},
Muhammad Umer\textsuperscript{\rm 1},
Abhiram Rao Gorle\textsuperscript{\rm 1},\\
Guijin Son\textsuperscript{\rm 2},
Youngjae Yu\textsuperscript{\rm 2},
John M. Cioffi\textsuperscript{\rm 1}
}
\affiliations{
\textsuperscript{\rm 1}Stanford University\\
\textsuperscript{\rm 2}Seoul National University
}

\begin{document}

\maketitle

\begin{abstract}
Small proxy models are commonly used to identify data mixtures for larger-scale training.
We ask whether their training trajectories reveal another transferable structure: the order in which larger models should resolve skill bottlenecks.
We formulate \textit{first-passage skill training}, where each monitored skill has a target floor and the objective is to minimize the tokens required to reach all floors.
We introduce \textbf{LogFloor}, a closed-loop controller that directs each round toward current bottlenecks, producing phase-ordered resolution trajectories.
Across five bAbI skill slices on Qwen2.5-1.5B, LogFloor reduces token cost by $56.2\%$ on average.
In 70M-to-12B transfer, three-round replay of a 70M scout path reaches every floor in all eight target runs, saving $30.9\%$ by pair mean, $39.4\%$ in pooled training tokens, and $37.6\%$ under source-cost accounting.
On MMLU-control, a frozen scout path succeeds across all eight 12B runs.
Collapsing a path to its static marginal mixture or reversing its phase order removes most benefits, while bottleneck labels alone remain partially useful.
These results identify phase-ordered bottleneck resolution as a transferable curriculum structure for monitored skill-targeted training.
\end{abstract}


\begin{figure*}[t]
  \centering
  \includegraphics[width=1\textwidth]{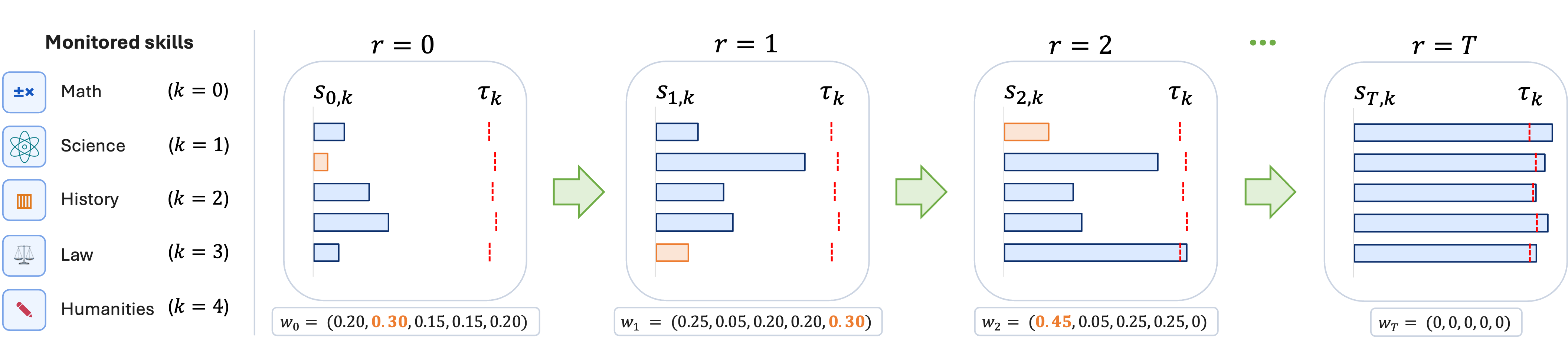}
  \caption{
    A simplified visualization of the proposed LogFloor controller for $K=5$ monitored skills.
    The bottleneck skill at each round $r$ is depicted in orange. At each round, \method{} shifts the largest share of the next allocation $w_r$ to this bottleneck, and the resulting ordered sequence $\{w_r\}$ forms the bottleneck-resolving trajectory that we later replay on larger target models.
  }
  \label{fig:overview}
\end{figure*}

\section{Introduction}

Data mixture selection is usually framed as a question of \emph{which} training data are most useful.
Small proxy models make this decision computationally tractable by scoring examples, estimating domain weights, or predicting large scale training outcomes.
Yet these approaches primarily seek useful mixtures or schedules under aggregate objectives.
We ask whether small-model trajectories reveal a different transferable structure: the order in which a larger model should resolve skill bottlenecks.

In many scenarios, the relevant objective is not a high average validation score across skills but satisfying a set of monitored skill thresholds, which we call floors.
For example, a model intended to support math, code, and multilingual use would not be suitable for use if even one skill remains below the required floor, regardless of its average score.
In such scenarios, training cannot terminate until the bottleneck skill reaches its floor.
The training policy must therefore determine which skills to prioritize at each stage of training.
We call this objective \emph{first-passage skill training}.
Given monitored skill slices and a target floor for each slice, the objective is to minimize the training tokens consumed until every skill reaches its floor.

Under this objective, uniform training with equal emphasis on each skill would waste resources on skills that have already reached their floors while a bottleneck skill has not yet reached its floor.
We introduce \textbf{\method}, a closed loop controller that observes per-skill monitor scores, computes floor pressure, and allocates the next training round toward the bottleneck skills exerting the largest pressure.
We call the resulting allocation path a \emph{bottleneck-resolving trajectory}.
Figure~\ref{fig:overview} provides a simplified illustration of how \method{} prioritizes the current bottleneck skill at each training round.

A 70M model can reveal a phase-ordered bottleneck path that remains useful for a 12B target after a short target side phase probe.
We do not claim that small models universally determine effective curricula, only that in monitored first-passage skill training, the transferable structure is a phase-ordered bottleneck-resolving trajectory and is not captured by a static mixture alone.

We support this claim with three key findings.
First, \method{} produces successful bottleneck-resolving trajectories.
On Qwen2.5-1.5B~\cite{qwen25} with five controlled bAbI skill slices~\cite{babi}, \method{} saves $56.2\%$ training tokens on average across target floors from $2.8$ to $3.3$.
Skill-It~\cite{skillit} and Online Data Mixing (ODM)~\cite{odm}, which do not use target information, do not robustly improve this first-passage objective.
Second, the path transfers across scale.
In the 70M to 12B setting, \emph{(replayable) target-probe replay} of a 70M scout trajectory reaches every target after three probe rounds, saving $30.9\%$ by pair mean, $39.4\%$ in pooled target training tokens, and $37.6\%$ under source cost accounting.
Third, the order matters.
On academic domain MMLU-control~\cite{mmlu}, a frozen 70M scout path hits all eight 12B route targets and saves $60.3\%$ in pooled rounds.
A clear scout decomposition shows that static mixture collapse and reversal in order preserve only $10.3\%$ and $6.0\%$, respectively.

The contributions of this paper are outlined below.
\begin{itemize}
    \item We formulate first-passage skill training, an objective that requires every monitored skill to reach its floor and thereby exposes bottleneck limited data control.
    \item We introduce \textbf{\method}, a simple floor pressure controller that produces bottleneck-resolving trajectories.
    \item We show that a 70M scout path can guide a 12B run via a short target-probe replay, retaining savings after accounting for the source controller cost.
    \item We isolate the transferred signal using order destruction controls, showing that bottleneck order is not reducible to a static data mixture.
\end{itemize}

\section{Related Work}

\subsection{Proxy Data Mixtures and Transfer}

A first line of work uses small proxy models to guide data selection and mixing for larger runs.
DoReMi~\cite{doremi} uses a small proxy model to select a useful static domain mixture for a larger training run.
SmallToLarge~\cite{s2l} uses small model training trajectories for example selection during supervised finetuning.
RegMix~\cite{regmix} uses regression over proxy runs to select a static mixture for larger scale training.
AC-ODM~\cite{acodm} transfers an actor trained with a proxy model to a larger target.
PROXYMIX~\cite{proxymix} transfers replay controllers learned on proxy models to larger targets.
Most recently, RegMix-D~\cite{regmixd} extends static proxy mixture selection to dynamic schedules learned from proxy loss trajectories.

We audit AC-ODM as a dynamic proxy-policy transfer baseline and RegMix as a static proxy-mixture baseline under the same monitored first-passage protocol.
RegMix-D is closest in using proxy trajectories for dynamic scheduling, but it evaluates schedules under a fixed training budget rather than minimizing the tokens required for all monitored skills to reach their floors.

\subsection{Online Data Control and Skill Acquisition Order}

A second line of work studies how data allocation and skill order evolve during training.
Curriculum and domain ordering work shows that the order of data can affect training trajectories~\cite{commute,curriculum}.
Within language model data control, Skill-It and ODM adapt data allocation online using skill losses and bandit feedback.
Aioli~\cite{aioli} later develops a unified optimization framework and an online method that updates mixture proportions during training.
Data mixing can also induce phase transitions that depend on the mixing ratio and model scale~\cite{phasetransition}.
TiKMiX~\cite{tikmix} uses group influence signals to track changing domain preferences and periodically update the data mixture.

Most recently, \citet{implicitcurriculum} analyze fixed threshold crossings across pretrained model checkpoints and show regularities in skill acquisition order across models.
Their work observes this order during ordinary pretraining, whereas we intervene on it by replaying a realized source path and test whether it reduces the first-passage training cost of a new target model.

\section{First-Passage Control with \method}

In this section, we formalize first-passage skill training and define the
\method{} controller.

\subsection{First-Passage Skill Training}

Let $K$ denote the number of monitored skills, indexed by $k=0,\ldots,K-1$, and let $\ell_{r,k}>0$ denote the average log loss on monitor set $k$ after training round $r$.
We measure the progress of skill $k$ relative to the model at round $0$ as
\begin{align}
    \label{eq:score}
    s_{r,k}=\frac{\ell_{0,k}}{\ell_{r,k}}.
\end{align}
Here, a score of $m$ indicates an $m$-fold reduction in monitored log loss.
Each skill has a target floor $\tau_k$, and the first-passage round $T$ is the first round at which all floors are reached:
\begin{align}
s_{T,k}\ge\tau_k \text{ } \text{for all } k.
\end{align}
Unlike fixed-budget objectives, this objective depends on whether every monitored skill reaches its floor, rather than on the average score at a fixed endpoint.
The scores $s_{r,k}$ determine when to stop and are used to update the controller, while the heldout split is used only as an external audit.
The bAbI experiments use the prespecified absolute floor sweep from $2.8$ to $3.3$. 
For MMLU-control, each target seed's floor is fixed before method evaluation as $90\%$ of the peak worst-skill route score in its $60$-round uniform trace.

We measure token saving relative to matched uniform training as
\begin{align}
\label{eq:savings}
1-\frac{C_{\mathrm{\method{}}}}{C_{\mathrm{uniform}}},
\end{align}
where each cost includes the training tokens consumed up to its first-passage round.
The main evaluation terms used throughout the paper are summarized in Table~\ref{tab:first-passage-terms}. 
\begin{table}[h]
\centering
\small
\setlength{\tabcolsep}{4pt}
\caption{\textbf{Evaluation terminology.}}
\label{tab:first-passage-terms}
\begin{tabular}{@{}p{0.30\linewidth}p{0.66\linewidth}@{}}
\toprule
Term & Definition \\
\midrule
Pooled saving
& Savings after summing token costs across runs. \\

Pair mean saving
& Mean of the savings of each matched run. \\

Hit
& All monitored floors are reached within the horizon. \\



Route first-hit round
& First round where all monitored floors are reached. \\

Heldout audit
& Heldout evaluation at the route first-hit round. \\

W/T/L
& Wins, ties, and losses against matched uniform training. \\
\bottomrule
\end{tabular}
\end{table}

\subsection{Floor-Pressure Control}

At each round $r$, the controller evaluates the monitor sets and computes
each skill's remaining deficit:
\begin{align}
    \label{eq:deficit}
    d_{r,k}=[\tau_k-s_{r,k}]^+,
\end{align}
where $[z]^+=\max\{z,0\}$.
The controller allocates more of the next training round to skills with
larger deficits.

Let $q_k$ be a fixed base share, let $\eta>0$ control how sharply the
controller focuses on the largest deficit, and let $\alpha\in[0,1]$ be the
smoothing coefficient.
Let $\operatorname{Bound}(\cdot)$ apply the fixed allocation caps and
renormalize the resulting vector.
The controller first forms a soft allocation and then smoothens it using the
previous allocation:
\begin{align}
    \label{eq:soft-allocation}
    \tilde w_{r,k}
    &=
    \frac{q_k\exp(d_{r,k}/\eta)}
         {\sum_j q_j\exp(d_{r,j}/\eta)},\\
    \label{eq:allocation}
    w_r
    &\leftarrow
    (1-\alpha)w_{r-1}
    +\alpha\,\operatorname{Bound}(\tilde w_r).
\end{align}
Here, $\tilde w_r = [{\tilde w}_{r,0}, {\tilde w}_{r,1}, \ldots, {\tilde w}_{r,K-1}]^\top$.
Thus, $\eta$ controls bottleneck emphasis, $\alpha$ controls how quickly
allocations change, and $\operatorname{Bound}(\cdot)$ enforces the fixed
allocation caps.
\method{} uses this soft exponential allocation.
We additionally evaluate Greedy \method{}, its hard-bottleneck variant.
Algorithm~\ref{alg:logfloor} summarizes the complete controller procedure,
including initialization, stopping, integer rounding, and trajectory
construction.
Full derivation is provided in the supplementary.

\begin{algorithm}[t]
\small
\caption{\method{} floor-pressure update}
\label{alg:logfloor}
\begin{algorithmic}[1]
\STATE \textbf{Input:} monitor sets; initial monitor losses $\ell_{0,k}$;
score floors $\tau_k$
\FOR{$r=0,1,\ldots$}
\STATE Compute scores $s_{r,k}$ using Eq.~\eqref{eq:score}
\STATE Compute deficits $d_{r,k}$ using Eq.~\eqref{eq:deficit}
\IF{$d_{r,k}=0$ for all $k$}
\STATE \textbf{break}
\ENDIF
\STATE Compute $\tilde w_r$ using Eq.~\eqref{eq:soft-allocation}
\IF{$r=0$}
\STATE Set $w_0=\operatorname{Bound}(\tilde w_0)$
\ELSE
\STATE Update $w_r$ using Eq.~\eqref{eq:allocation}
\ENDIF
\STATE Train one round according to $w_r$ using largest-remainder rounding
\ENDFOR
\STATE \textbf{Output:} ordered allocation trajectory
$\gamma=\{w_r\}_{r=0}^{T-1}$
\end{algorithmic}
\end{algorithm}
\subsection{First-Passage Efficiency}

\begin{figure}[t]
\centering
\includegraphics[width=0.9\linewidth]{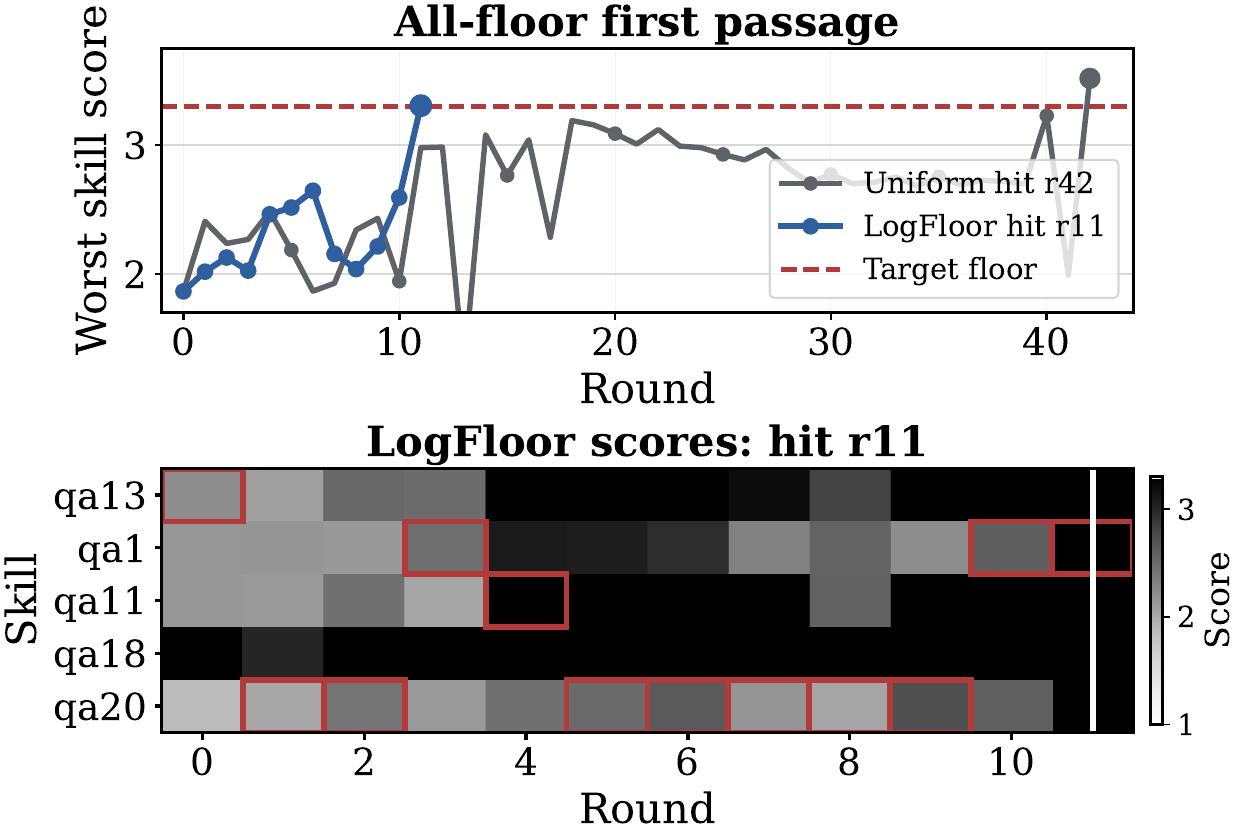}
\caption{\textbf{A representative bottleneck-resolving trajectory.}
Uniform training clears all target floors at zero-indexed round 42, while \method{} clears them at zero-indexed round 11 by repeatedly prioritizing this floor-pressure bottleneck. Red boxes in the lower heatmap identify the bottleneck skill selected at each round.}
\label{fig:trajectory-mechanism}
\end{figure}

Before testing the transfer to a larger model, we first show that \method{} produces successful first-passage trajectories.
In the representative run in Figure~\ref{fig:trajectory-mechanism}, uniform training clears all target floors at zero-indexed round $42$, whereas \method{} reaches them at round $11$ by repeatedly prioritizing the current floor-pressure bottleneck skill.
The red boxes in the lower panel mark the bottleneck skill selected at each round.

Table~\ref{tab:qwen-target-sweep-main} summarizes the target sweep on Qwen2.5-1.5B over five controlled bAbI skill slices.
Across six absolute target floors from $2.8$ to $3.3$ and five random seeds per floor, \method{} achieves $56.2\%$ pooled token savings relative to matched uniform runs.
It reaches all $30$ route targets, with an exact $95\%$ Clopper--Pearson interval of $[88.4\%,100\%]$ for the route-hit rate.
Greedy \method{} also performs well with $51.1\%$ pooled savings, supporting the intuition that floor pressure is the dominant indicator of allocation.
\method{} wins in $20$ out of $30$ cases against the greedy variant, indicating greater stability across seed and target pairs.
Target-free online data-mixing baselines perform poorly for this first-passage objective, with pooled savings of $-7.8\%$ for Skill-It and $-0.9\%$ for ODM.

We now examine heldout performance.
At the route first-hit round, the heldout audit clears all target floors in $28$ out of $30$ runs.
The mean difference between the heldout and route worst-skill scores is $+0.049$, while the mean minimum heldout margin above the corresponding target floors is $+0.260$.
The exact $95\%$ Clopper--Pearson interval for the heldout confirmation rate is $[77.9\%,99.2\%]$.
Thus, route first-passage is usually corroborated on heldout data.

Regarding the experimental setup, all compared methods use the same five slices, route/heldout split, batch size, allocation bounds, monitor cadence, stopping rule, and horizon for each seed and floor.
Controller hyperparameters are fixed in advance, and Skill-It and ODM do not use target information in their allocation updates.

\begin{table}[h]
\centering
\small
\setlength{\tabcolsep}{3pt}
\caption{\textbf{First-passage target sweep on Qwen2.5-1.5B.}
Each target floor has five seeds.
\method{} reaches the route target in $30/30$ runs, and $28/30$ route first-hits also clear the heldout target on all floors.}
\label{tab:qwen-target-sweep-main}
\begin{tabular}{@{}c r r r r r@{}}
\toprule
Floor & \shortstack{\method{}\\pair mean} & \shortstack{\method{}\\pooled} & \shortstack{Greedy\\pooled} & \shortstack{Skill-It\\pooled} & \shortstack{ODM\\pooled} \\
\midrule
2.8 & 36.3\% & 59.2\% & 54.4\% & -12.1\% & 7.2\% \\
2.9 & 28.0\% & 53.2\% & 52.0\% & -22.9\% & 4.8\% \\
3.0 & 27.1\% & 49.1\% & 49.1\% & -13.3\% & 12.2\% \\
3.1 & 48.7\% & 55.6\% & 55.7\% & 3.6\% & 20.9\% \\
3.2 & 57.8\% & 62.2\% & 52.4\% & -5.3\% & -19.6\% \\
3.3 & 45.0\% & 55.4\% & 44.9\% & -3.8\% & -17.9\% \\
\midrule
All floors & 40.5\% & 56.2\% & 51.1\% & -7.8\% & -0.9\% \\
\bottomrule
\end{tabular}
\end{table}

These results motivate using \method{} also as a trajectory generator rather than only as a per-run controller.
Its output is interpretable as an ordered sequence of bottleneck decisions, which is the object we test for transfer across scale.
\section{Cross-Scale Transfer by Replaying \method{} Trajectories}

We evaluate whether bottleneck-resolving trajectories learned by small models
can guide larger target models.

\subsection{Trajectory Replay Schemes}

For a source run $S$, we define the transferable object as its ordered
before-hit allocation sequence:
\begin{align}
    \gamma^S=\{w_0^S,w_1^S,\ldots,w_{T_S-1}^S\},
\end{align}
where $T_S$ is the first-passage round of the source.
Since source and target models may traverse bottleneck phases at different
rates, a replay scheme must map each target round to an allocation in the
source trajectory.
We refer to a monotone traversal of the ordered source path after phase
localization as a phase-indexed replay.

\paragraph{Raw round-indexed replay.}
This scheme directly applies the source allocation $w_r^S$ at target round
$r$.
It assumes that the source and target traverse their bottleneck phases at the
same speed, which is generally false.
We use it as a baseline to test whether literal source-round copying is
sufficient without phase localization.

\paragraph{(Replayable) target-probe replay.}
This scheme uses a short uniform target probe to localize the target to a source phase and then advances monotonically through the ordered source trajectory.
Let $H$ denote the probe length.
We use a single global probe length of $H=3$ in all main-result target-probe experiments.
We chose it as a common operating point after preliminary $H\in\{1,3,5\}$
sensitivity checks. In the MMLU source-seed-2 sweep, $H=3$ reaches all $8/8$
targets with $31.0\%$ pooled-round saving, versus $7/8$ for $H=1$ and
$21.6\%$ for $H=5$; $H$ is not retuned by target seed, floor, or
source--target pair.
The Supplementary Document provides the exact protocol and the sensitivity analysis for $H \in \{1, 3, 5\}$. 

The resulting route-score window is normalized and matched to same-length normalized windows from a 70M uniform calibration trace.
We slide the target's three-round probe pattern along the 70M calibration trace and choose the source phase that looks the most similar. 
Similarity is measured from the relative skill scores within each round and the trend of the worst skill, using normalized mean squared distance. 
Let $j^\star$ denote the start of the matched source window.
For target round $r\ge H$, the replayed source-action index is
\begin{align}
    \label{eq:target-probe-replay}
    i_r=\min\{j^\star+r-H,\,T_S-1\}.
\end{align}
The target applies $w_{i_r}^S$, advancing monotonically from the matched source phase and clamping to the final logged allocation if it outlasts the source trajectory.
After the probe, replay uses no further target scores or target first-passage horizon to update, interpolate, or rematch the schedule.
We use this scheme to test whether a small-model trajectory can guide a larger target after only short phase localization and without continued target-side adaptation.

\paragraph{(Oracle) phase-aligned replay.}
This scheme length-normalizes the source trajectory using the observed before-hit target length $T_L$.
At target round $r$, it applies the source allocation nearest to the same relative before-hit progress $r/(T_L-1)$.
Since $T_L$ is available only after observing the target trajectory, this scheme is not replayable at deployment.
We use it as an oracle diagnostic to test whether the source trajectory remains useful when the source and target are placed at matched bottleneck phases.

\subsection{70M-to-12B Transfer}

\begin{figure}[t]
    \centering
    \includegraphics[width=0.8\linewidth]{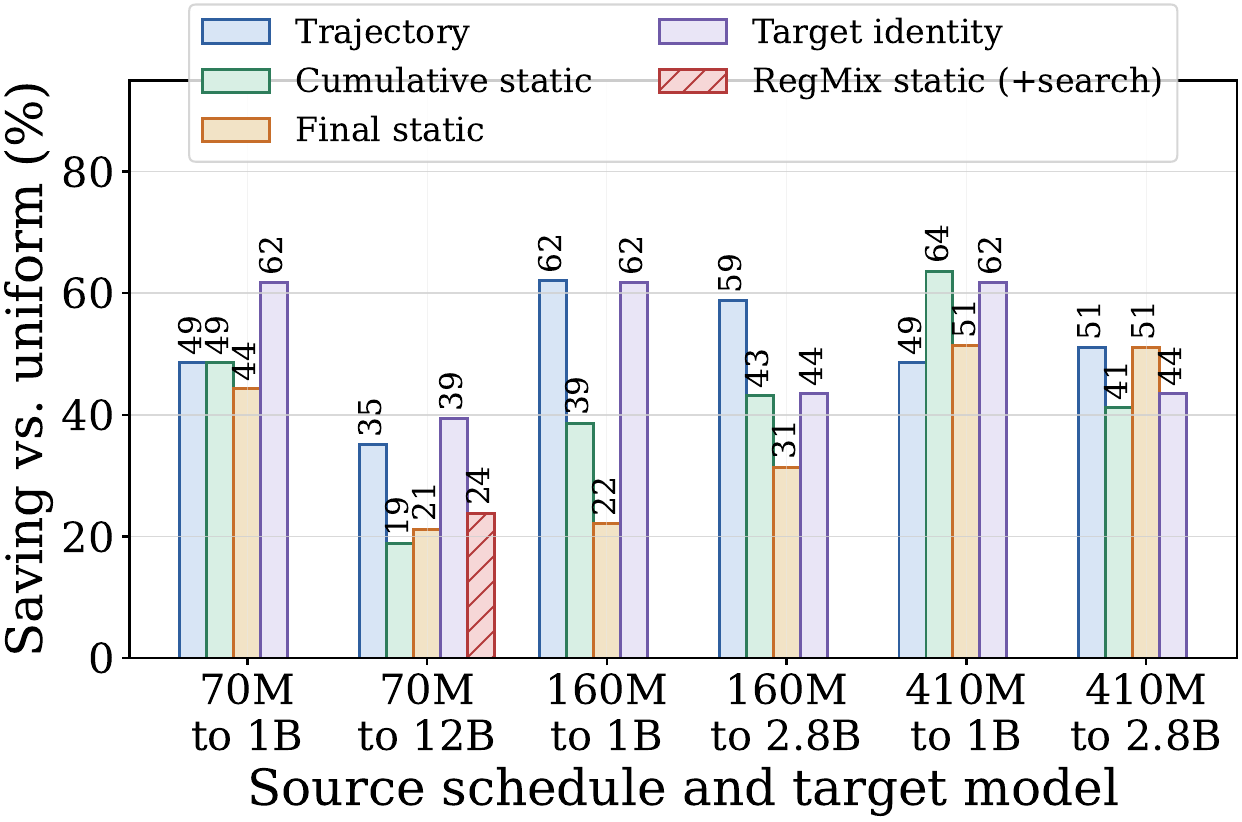}
    \caption{\textbf{Small-to-large trajectory transfer.} Replayable target-probe replay transfers source \method{} trajectories across six Pythia settings and outperforms static summaries on average.}
    \label{fig:small-to-large-transfer}
\end{figure}

We first test whether \emph{(replayable) target-probe replay} transfers source trajectories across models with different scales.
Six Pythia~\cite{pythia} transfer settings using replayable target-probe replay are summarized in Figure~\ref{fig:small-to-large-transfer}.
In these settings, the trajectory replay achieves an unweighted mean target training token savings of $51.0\%$, compared to $42.3\%$ for cumulative static mixtures and $36.9\%$ for final static mixtures.
Thus, small-to-large trajectory transfer works across multiple scale pairs
and preserves more benefit on average than both static mixture baselines.
70M-to-12B achieves the largest savings, where a path discovered by the smallest
Pythia model remains useful across a roughly $170\times$ increase in model size.

Table~\ref{tab:transfer-aggregate-main} focuses on the 70M-to-12B setting.
These bAbI rows use eight seed-matched 70M--12B source--target pairs.
The 12B online \method{} row is the fully target-adaptive reference, the
frozen target mixture isolates the effect of the probe, and phase-aligned
replay is an oracle diagnostic.
Table~\ref{tab:transfer-aggregate-main} gives us three main insights.
First, replaying the 70M trajectory is nearly as effective as running \method{} online at 12B scale.
The 12B online controller saves $34.4\%$ by pair mean and $39.4\%$ pooled savings, while the replayable $3$-round target-probe method saves $30.9\%$ by pair mean and $39.4\%$ pooled, reaching all eight targets and winning seven out of eight matched comparisons.
Paired bootstrap resampling of the eight target seeds gives a $95\%$ interval of $[4.6\%,50.9\%]$ for the target-probe pair-mean saving.

Second, the transferred source trajectory provides value beyond the $3$-round target probe.
A target mixture computed from the same probe and kept constant throughout the remaining training reaches only seven out of eight targets.
The remaining seed does not hit within the horizon, so its aggregate cost and savings are undefined.
In contrast, replaying the 70M trajectory reaches all eight targets, showing that the ordered source path provides useful information beyond the static mixture obtained from the probe.
As an oracle diagnostic, the phase-aligned replay also reaches all eight targets and saves $35.1\%$ pooled, confirming that the source trajectory remains useful when its phase is correctly aligned with the target.

\begin{figure}[t]
    \centering
    \includegraphics[width=0.98\linewidth]{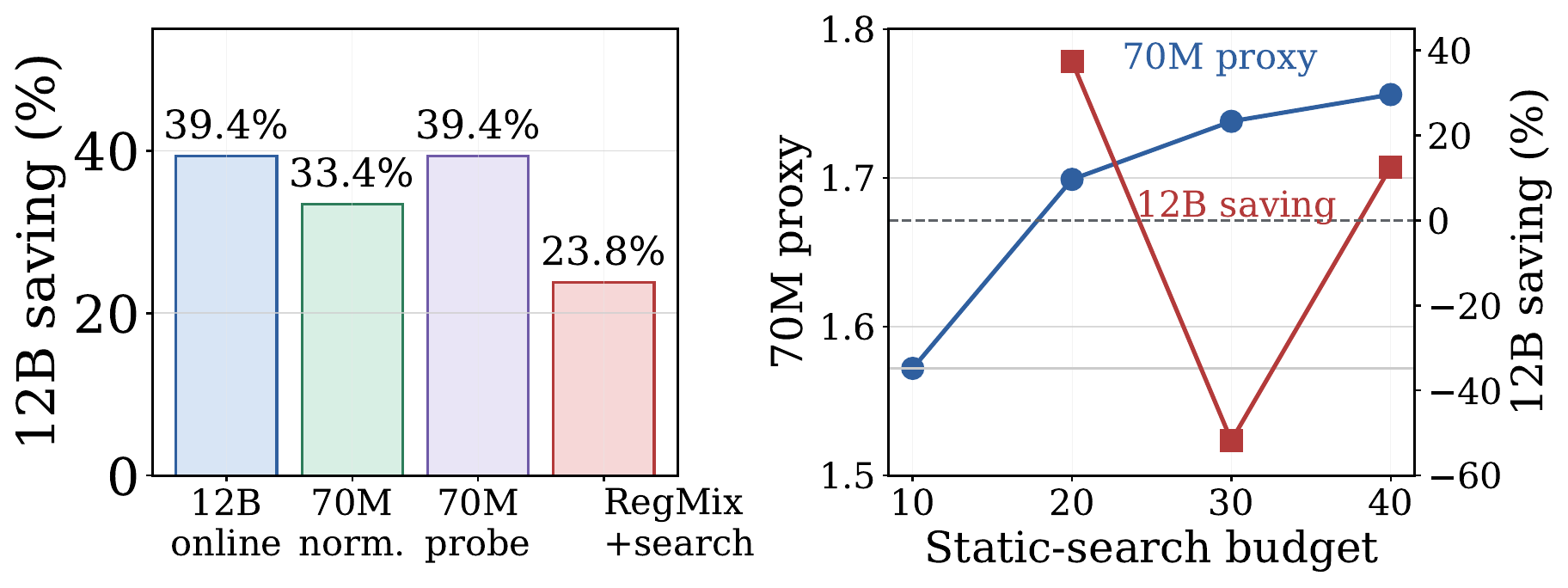}
    \caption{\textbf{70M-to-12B transfer diagnostics.}
    The left plot juxtaposes target-only savings for 12B online and 70M target-probe replay with cost-inclusive savings for 70M phase replay and RegMix.
    The right plot shows that static proxy search improves the 70M proxy score but transfers non-monotonically to 12B.}
    \label{fig:transfer-diagnostics}
\end{figure}

Third, replay remains useful even after accounting for the cost of scouting the trajectory.
We multiply the logged 70M source training tokens by the parameter ratio $70\mathrm{M}/12\mathrm{B}$ and add the resulting cost equivalent to 12B to the replay target cost.
Nonetheless, replay still saves $37.6\%$.
Figure~\ref{fig:transfer-diagnostics} (left) compares target-only with cost-inclusive savings and Table~\ref{tab:transfer-aggregate-main} gives target-probe replay's $37.6\%$ scout-inclusive saving.

\begin{table}[t]
\centering
\small
\setlength{\tabcolsep}{2pt}
\caption{\textbf{70M-to-12B transfer accounting.} Replayability, probe, oracle, and source-cost controls over eight target runs. Hit and Win count route hits and wins against matched uniform runs; rows with a missed target omit aggregate savings.}
\label{tab:transfer-aggregate-main}
\begin{tabular}{@{}>{\raggedright\arraybackslash}p{0.32\linewidth}ccrrrr@{}}
\toprule
Method & Hit & Win & \shortstack[r]{Target\\tokens} & \shortstack[r]{Pair\\mean} & Pooled & \shortstack[r]{Incl.\\scout} \\
\midrule
12B uniform & 8/8 & -- & 825{,}094 & 0.0\% & 0.0\% & 0.0\% \\
12B online \method{} & 8/8 & 8/8 & 499{,}711 & 34.4\% & 39.4\% & 39.4\% \\
70M target-probe replay & 8/8 & 7/8 & 500{,}375 & 30.9\% & 39.4\% & 37.6\% \\
Probe + frozen target mix & 7/8 & -- & -- & -- & -- & -- \\
70M phase replay (oracle) & 8/8 & 7/8 & 535{,}150 & 30.1\% & 35.1\% & 33.4\% \\
\bottomrule
\end{tabular}
\end{table}

Finally, we compare replay with proxy-based baselines.
On the matched three-seed ($n=3$) subset, RegMix saves $40.1\%$ when only target training is counted, but $23.8\%$ after including its 70M proxy-search cost.
The trajectory replay saves $35.3\%$ under the same accounting.
Figure~\ref{fig:transfer-diagnostics} (right) exposes a limitation of static proxy search.
As the search budget increases, the selected mixture improves on the 70M proxy, but its transferred performance on 12B is non-monotonic.
Thus, the mixture ranked best by the small model need not be the mixture that is best for the large model.
The AC-ODM proxy actor reaches all three targets but uses 216K target training tokens, compared to 199K for replay.
Thus, replay is more efficient under cost-aware comparison.
The Supplementary Document provides the RegMix audit and shows that replay should first match the target's state to a source phase. Directly copying source allocations by round can fail.

\section{Understanding Replay-Based Transfer}

In this section, we examine whether replay transfers phase-ordered bottleneck
information rather than a static mixture or a seed-paired trace.

\subsection{Shared Bottleneck Geometry Across Scale}

Phase-indexed replay is plausible if source and target models encounter similar bottleneck states at the corresponding stages of training.
Let $\phi\in[0,1]$ denote the relative progress towards the first-passage, so that $\phi=0$ is the beginning of the trajectory and $\phi=1$ is the first-hit boundary.
This allows us to compare source and target models at the same normalized phase rather than at the same training round.
Across normalized training phase, the aligned 70M source trajectory and the 12B online controller emphasize largely overlapping bottlenecks.

For each Pythia-70M source and Pythia-12B target online \method{} run, we truncate both trajectories before the first hit, align them by $\phi$, and compare their deficit vectors, controller allocation vectors, and top bottleneck identities.
Table~\ref{tab:phase-geometry-main} shows that off-diagonal source--target seed pairs are as similar as diagonal pairs.
The deficit cosine is $0.807$ for diagonal pairs and $0.814$ for off-diagonal pairs, while the overlaps of the top-$2$ bottlenecks are $0.930$ and $0.935$, respectively.
The source trajectory is therefore not merely a trace specific to one matched source--target seed pair.

\begin{table}[t]
\centering
\small
\setlength{\tabcolsep}{2pt}
\caption{\textbf{70M-to-12B phase geometry audit.} Before-hit trajectories are aligned by normalized phase, and entries are means over diagonal and off-diagonal source--target seed pairs.}
\label{tab:phase-geometry-main}
\begin{tabular}{@{}p{0.35\linewidth}rccc@{}}
\toprule
Comparison & Pairs & Deficit cos & Alloc. cos & Top-$2$ \\
\midrule
Diagonal seed $i\!\to\!i$ & $8$ & $0.807$ & $0.844$ & $0.930$ \\
Off-diagonal $i\!\to\!j$ & $56$ & $0.814$ & $0.839$ & $0.935$ \\
\bottomrule
\end{tabular}
\end{table}

The shared deficit geometry also has a simple local interpretation.
Let $d_\phi^S$ and $d_\phi^L$ denote the source and target deficit vectors at the same aligned phase, and let $\tilde w_\phi^S$ and $\tilde w_\phi^L$ be their soft allocations from Eq.~\eqref{eq:soft-allocation}.
For fixed $q$ and $\eta$, these allocations satisfy
\begin{align}
\label{eq:matched-phase-continuity}
\left\|\tilde w_\phi^S-\tilde w_\phi^L\right\|_1
\le
\frac{1}{\eta}
\left\|d_\phi^S-d_\phi^L\right\|_\infty.
\end{align}
This follows because Eq.~\eqref{eq:soft-allocation} is a softmax in $d/\eta$, and its sensitivity to changes in $d$ is bounded by $1/\eta$. Applying this bound between the source and target deficits gives Eq.~\eqref{eq:matched-phase-continuity}.
Thus, similar deficits at a matched phase imply similar ideal soft
allocations.

This is only a local continuity statement.
It does not model the full training dynamics or prove first-passage transfer, and the executed allocation additionally includes bounding, smoothing, and integer rounding.
Nevertheless, in the MMLU matched-target local audit, the logged source actions retain $98.5\%$ of the target-online pressure progress, with a mean one-step surrogate gap of $0.0236$.
The full derivation and the corresponding implementation-gap analysis are provided in the Supplementary Document.

\subsection{Order, Not Static Mixture}

This subsection asks what kind of information in the source trajectory can be transferred.
Table~\ref{tab:mmlu-transfer-compact} reports the matched and fixed-source replay diagnostics.
The order controls summarized in Table~\ref{tab:mmlu-order-destruction-main} then show whether the gain comes from temporal order rather than the static mixture, and how much of that gain is retained by the bottleneck sequence alone.

We use MMLU control because its five academic domains are easily interpretable and allow controlled interventions on the same frozen scout path.
The domains are math, science, history, law/social science, and humanities.
This is a controller diagnostic with separate route and heldout splits, not a
leaderboard MMLU evaluation.

\paragraph{Transfer across seeds.}
Table~\ref{tab:mmlu-transfer-compact} shows that a single frozen 70M scout can be replayed across all eight 12B target seeds.
The fixed source seed $2$ reaches all eight targets and saves $60.3\%$ in pooled rounds.
The order diagnostic extends beyond source seed $2$.
Across source seeds $0$--$2$, correctly ordered replay reaches all $24$
target runs, whereas every static or reversed control is either less efficient
than its matched ordered path or contains at least one no-hit
(Supplementary Document).
Matched replay saves $38.8\%$, compared to $33.6\%$ for 12B online
\method{}.
The fixed target-floor protocol and seed-level audit are provided in the Supplementary Document.

The three-round target probe replay also reaches all eight targets and saves $31.0\%$ in pooled rounds.
By contrast, freezing the target mixture obtained after the same probe reaches only seven out of eight targets.
Thus, the replay result is not explained by the probe alone. 

\begin{table}[h]
\centering
\small
\setlength{\tabcolsep}{2pt}
\caption{\textbf{MMLU control transfer diagnostics.}
Rows aggregate eight Pythia-12B target seeds unless noted.
W/T/L denotes wins, ties, and losses against matched uniform runs.
Rows with a missed target omit aggregate savings.}
\label{tab:mmlu-transfer-compact}
\begin{tabular}{@{}p{0.50\linewidth}cccc@{}}
\toprule
Schedule & Hit & W/T/L & \shortstack{Pair\\mean} &
\shortstack{Pooled\\rounds} \\
\midrule
12B online \method{} & $8/8$ & $4/3/1$ & $25.6\%$ & $33.6\%$ \\
70M matched replay & $8/8$ & $5/1/2$ & $23.5\%$ & $38.8\%$ \\
Frozen scout seed $2$ & $8/8$ & $6/2/0$ & $46.8\%$ & $60.3\%$ \\
$3$-round target probe & $8/8$ & $5/2/1$ & $17.2\%$ & $31.0\%$ \\
$3$-round probe + frozen target mix & $7/8$ & $4/1/3$ & -- & -- \\
\bottomrule
\end{tabular}
\end{table}

\paragraph{Order and retained information.}
The order controls in Table~\ref{tab:mmlu-order-destruction-main} decompose the information in the frozen trajectory.
We fix source seed $2$ across the eight target seeds and all order interventions.
Source seed $2$ was selected before any 12B runs using only source-side
route/heldout agreement and controller diagnostics. It is the headline
decomposition because its correct, static, reverse, and label controls all
hit $8/8$, making their pooled comparisons defined: correct replay saves
$60.3\%$ in pooled rounds, versus $10.3\%$ for static collapse and $6.0\%$
for reversal. The source-0 and source-1 ordered paths also reach all $16/16$
targets, but their corresponding control suites contain no-hit rows and are
reported in the Supplementary Document.

\begin{table}[h]
\centering
\small
\setlength{\tabcolsep}{2pt}
\caption{\textbf{Decomposition of the frozen source seed 2 trajectory.}
All rows use the same eight Pythia-12B targets and the same floors for each
target.
Rows with a missed target omit aggregate savings.}
\label{tab:mmlu-order-destruction-main}
\resizebox{\linewidth}{!}{%
\begin{tabular}{@{}p{0.25\linewidth}p{0.29\linewidth}cccc@{}}
\toprule
Schedule & Retained signal & Hit & \shortstack{Pair\\mean} &
\shortstack{Pooled\\rounds} & \shortstack{Pooled\\tokens} \\
\midrule
Correct seed $2$ scout
& full phase indexed allocation path
& $8/8$ & $46.8\%$ & $60.3\%$ & $59.8\%$ \\
Static collapse
& marginal allocation only
& $8/8$ & $-2.9\%$ & $10.3\%$ & $9.1\%$ \\
Reverse
& same allocations in reverse order
& $8/8$ & $-13.6\%$ & $6.0\%$ & $5.3\%$ \\
Random permutation
& same allocations in shuffled order
& $6/8$ & -- & -- & -- \\
Label replay
& bottleneck identities in correct order
& $8/8$ & $27.9\%$ & $25.0\%$ & $22.4\%$ \\
Label reverse
& same identities in reverse order
& $8/8$ & $-66.7\%$ & $-11.2\%$ & $-10.8\%$ \\
\bottomrule
\end{tabular}
}
\end{table}

The correct path saves $46.8\%$ by pair mean, $60.3\%$ in pooled rounds, and $59.8\%$ in pooled tokens.
Collapsing the path to its static marginal mixture reduces the pooled round savings to $10.3\%$.
Reversing the same allocation vectors reduces it to $6.0\%$, and a random permutation reaches only six out of eight targets.
Across the same eight target seeds, the correctly ordered path is faster than static collapse in $6/8$ cases, ties once, and is slower once, using $58$ fewer first-passage rounds in total; the corresponding comparison with reversal is also $6/1/1$, with $63$ fewer total rounds.
These controls preserve either the same allocation vectors or their marginal mixture while changing the temporal structure.
The resulting loss therefore shows that order carries most of the transfer gain.

Keeping only the bottleneck identities in the correct order still saves $25.0\%$ in pooled rounds, while reversing the same sequence gives $-11.2\%$.
The bottleneck sequence therefore carries useful coarse information.
The $60.3\%$ savings from the full path shows that allocation strength and secondary domain structure provide substantial additional value.

A separate bAbI replication gives the same qualitative result.
Correct replay reaches all eight targets and saves $35.1\%$ in pooled tokens, static mean replay saves $20.1\%$, and reversal reaches only six targets with $-1.9\%$ savings.
Mass assigned to the current bottleneck correlates positively with token saving ($\rho=0.589$), while mass assigned to already cleared skills correlates negatively ($\rho=-0.704$).

\section{Robustness, Scope, and Limitations}

\subsection{Replay Across Target Floors}

Replay does not need to be scouted at the exact deployment floor.
We replay Qwen2.5-1.5B \method{} trajectories scouted at floors $2.8$, $3.0$, and $3.3$ at the other two target floors after the same three round target probe.
Table~\ref{tab:cross-floor-replay} reports the six different pairs of scout and target
floors over five target seeds each.
The replay reaches $28$ out of $30$ targets, and four out of the six pairs reach all five.
Most importantly, the replay succeeds when the scout floor is either lower or higher than the target floor, showing that exact floor agreement is unnecessary within the evaluated range.

\begin{table}[t]
\centering
\small
\caption{\textbf{Replay across target floors on Qwen2.5-1.5B.}
Each cell gives route hits out of five target seeds after the same three-round target probe; darker shading marks full $5/5$ hits.
Gray cells mark identical scout and target floors.}
\label{tab:cross-floor-replay}
\begin{tabular}{@{}cccc@{}}
\toprule
 & \multicolumn{3}{c}{Target floor} \\
\cmidrule(lr){2-4}
Scout floor & $2.8$ & $3.0$ & $3.3$ \\
\midrule
$2.8$ & \cellcolor{black!8}-- & \cellcolor{teal!12}$4/5$ & \cellcolor{teal!30}$5/5$ \\
$3.0$ & \cellcolor{teal!30}$5/5$ & \cellcolor{black!8}-- & \cellcolor{teal!12}$4/5$ \\
$3.3$ & \cellcolor{teal!30}$5/5$ & \cellcolor{teal!30}$5/5$ & \cellcolor{black!8}-- \\
\bottomrule
\end{tabular}
\end{table}

\subsection{Robustness and Scope}

Table~\ref{tab:breadth-main} summarizes robustness across target levels,
skill counts, model sizes, pretraining stages, benchmarks, and architectures.
Detailed results, seed level tables, and censoring audits are provided in the
Supplementary Document.

\begin{table}[t]
\centering
\small
\setlength{\tabcolsep}{3pt}
\caption{\textbf{Breadth and scope summary.}
}
\label{tab:breadth-main}
\begin{tabular}{@{}>{\raggedright\arraybackslash}p{0.23\linewidth}>{\raggedright\arraybackslash}p{0.47\linewidth}>{\raggedright\arraybackslash}p{0.21\linewidth}@{}}
\toprule
Axis & Result & Role \\
\midrule
Target sweep & Qwen2.5-1.5B saves $56.2\%$ pooled tokens across six floors & primary validation \\
\addlinespace
Skill count & $K\in\{5,7,10,15\}$: $41.8\%$ aggregate and $40.1\%$ paired savings & floor-count robustness \\
\addlinespace
Model size & Pythia 70M--2.8B wins on all evaluated seeds, $42.1\%$ mean saving & scale robustness \\
\addlinespace
Pretraining stage & positive from step 16K onward, $53.9\%$ at the final checkpoint & stage robustness \\
\addlinespace
SuperGLUE control & $51.2\%$ (base) and $53.7\%$ (CB-substituted) route savings; $7/10$ heldout-confirmed & beyond bAbI \\
\addlinespace
Architecture and frontier & Mamba self-control positive; single-seed 72B pilot hits at round 7 (uniform), 2 (online), 3 (replay) & scope evidence \\
\bottomrule
\end{tabular}
\end{table}

Note that the skill count and model size results establish robustness, not
monotonic improvement with either $K$ or parameter count.
The gain is weak at the earliest pretraining stage but becomes consistently
positive at later checkpoints.
The SuperGLUE, Mamba, and 72B results further extend the evidence beyond the
main bAbI and Pythia settings.
Together, these results define the empirical scope of the method rather than
a universal curriculum law.

\subsection{Limitations}

The target side online \method{} and the small model replay are two operating modes of the same control principle, not competing methods.
Our evidence concerns monitored first-passage training with explicit floors and data slices associated with each skill.
The target probe replay is deployable, whereas the oracle phase-aligned replay is used only as a mechanism diagnostic.
Replay across architectures is not guaranteed. 
For instance, a replay from Qwen to Mamba transfers well, whereas a replay from Mamba to Qwen gives mixed results.
The controller also treats monitored skills separately and does not explicitly model transfer or interference among them.

\section{Conclusion}

We introduced first-passage skill training and \method, a floor-pressure controller that exposes bottleneck-resolving trajectories.
The key takeaway is that these trajectories can be transferred across scale. 
A 70M model can scout a phase-ordered bottleneck path that helps a 12B target reach all monitored floors faster after a short target probe.
In bAbI-control, this replay nearly matches the direct 12B online control. 
In the MMLU-control order diagnostic, frozen 70M scout paths can even give a stronger pooled round signal.
Order destruction controls show that the reusable signal is not merely a static data mixture.
Reversing or collapsing the path removes most of the gain, while the bottleneck label sequence remains partially useful.
Small models therefore do not only estimate useful mixtures.
In monitored first-passage training, they can reveal the order in which larger models should resolve skill bottlenecks.

\bibliography{aaai2027}

@article{doremi,
  title={Doremi: Optimizing data mixtures speeds up language model pretraining},
  author={Xie, Sang Michael and Pham, Hieu and Dong, Xuanyi and Du, Nan and Liu, Hanxiao and Lu, Yifeng and Liang, Percy S and Le, Quoc V and Ma, Tengyu and Yu, Adams Wei},
  journal={Advances in Neural Information Processing Systems},
  volume={36},
  pages={69798--69818},
  year={2023}
}

@article{skillit,
  title={Skill-it! a data-driven skills framework for understanding and training language models},
  author={Chen, Mayee and Roberts, Nicholas and Bhatia, Kush and Wang, Jue and Zhang, Ce and Sala, Frederic and R{\'e}, Christopher},
  journal={Advances in Neural Information Processing Systems},
  volume={36},
  pages={36000--36040},
  year={2023}
}

@article{odm,
  title={Efficient online data mixing for language model pre-training},
  author={Albalak, Alon and Pan, Liangming and Raffel, Colin and Wang, William Yang},
  journal={arXiv preprint arXiv:2312.02406},
  year={2023}
}

@inproceedings{regmix,
  title={Regmix: Data mixture as regression for language model pre-training},
  author={Liu, Qian and Zheng, Xiaosen and Muennighoff, Niklas and Zeng, Guangtao and Dou, Longxu and Pang, Tianyu and Jiang, Jing and Lin, Min},
  booktitle={International Conference on Learning Representations},
  volume={2025},
  pages={38305--38339},
  year={2025}
}

@article{regmixd,
  title={RegMix-D: Dynamic Data Mixing via Proxy Training Trajectories},
  author={Zhao, Kaiyan and Miao, Zhongtao and Aizawa, Akiko and Tsuruoka, Yoshimasa},
  journal={arXiv preprint arXiv:2606.18663},
  year={2026}
}

@article{s2l,
  title={Smalltolarge (s2l): Scalable data selection for fine-tuning large language models by summarizing training trajectories of small models},
  author={Yang, Yu and Mishra, Siddhartha and Chiang, Jeffrey and Mirzasoleiman, Baharan},
  journal={Advances in Neural Information Processing Systems},
  volume={37},
  pages={83465--83496},
  year={2024}
}

@article{proxymix,
  title={Dynamic Proxy-Mixing: Transferring Replay Controllers from Small to Large Models for Continual Instruction Tuning},
  author={Shihab, Ibne Farabi and Afrin, Fariya and Sharma, Anuj},
  journal={arXiv preprint arXiv:2606.00400},
  year={2026}
}

@misc{acodm,
      title={AC-ODM: Actor--Critic Online Data Mixing for Sample-Efficient LLM Pretraining}, 
      author={Jing Ma and Chenhao Dang and Mingjie Liao},
      year={2026},
      eprint={2505.23878},
      archivePrefix={arXiv},
      primaryClass={cs.LG},
      url={https://arxiv.org/abs/2505.23878}, 
}

@inproceedings{aioli,
  title={Aioli: A unified optimization framework for language model data mixing},
  author={Chen, Mayee and Hu, Michael and Lourie, Nicholas and Cho, Kyunghyun and R{\'e}, Christopher},
  booktitle={International Conference on Learning Representations},
  volume={2025},
  pages={46089--46132},
  year={2025}
}

@article{commute,
  title={Commute Your Domains: Trajectory Optimality Criterion for Multi-Domain Learning},
  author={Rukhovich, Alexey and Podolskiy, Alexander and Piontkovskaya, Irina},
  journal={arXiv preprint arXiv:2501.15556},
  year={2025}
}

@article{tikmix,
  title={TiKMiX: Take Data Influence into Dynamic Mixture for Language Model Pre-training},
  author={Wang, Yifan and Liu, Binbin and Liu, Fengze and Guo, Yuanfan and Deng, Jiyao and Wu, Xuecheng and Zhou, Weidong and Zhou, Xiaohuan and Wang, Taifeng},
  journal={arXiv preprint arXiv:2508.17677},
  year={2025}
}

@article{curriculum,
  title={Curriculum Learning for LLM Pretraining: An Analysis of Learning Dynamics},
  author={Elgaar, Mohamed and Amiri, Hadi},
  journal={arXiv preprint arXiv:2601.21698},
  year={2026}
}

@article{implicitcurriculum,
  title={What do language models learn and when? The implicit curriculum hypothesis},
  author={Liu, Emmy and Sun, Kaiser and Li, Millicent and Lee, Isabelle and Tjuatja, Lindia and Huang, Jen-tse and Neubig, Graham},
  journal={arXiv preprint arXiv:2604.08510},
  year={2026}
}

@article{phasetransition,
  title={Data mixing can induce phase transitions in knowledge acquisition},
  author={Gu, Xinran and Lyu, Kaifeng and Li, Jiazheng and Zhang, Jingzhao},
  journal={Advances in Neural Information Processing Systems},
  volume={38},
  pages={163200--163244},
  year={2026}
}

@article{babi,
  title={Towards ai-complete question answering: A set of prerequisite toy tasks},
  author={Weston, Jason and Bordes, Antoine and Chopra, Sumit and Rush, Alexander M and Van Merri{\"e}nboer, Bart and Joulin, Armand and Mikolov, Tomas},
  journal={arXiv preprint arXiv:1502.05698},
  year={2015}
}

@article{mmlu,
  title={Measuring massive multitask language understanding},
  author={Hendrycks, Dan and Burns, Collin and Basart, Steven and Zou, Andy and Mazeika, Mantas and Song, Dawn and Steinhardt, Jacob},
  journal={arXiv preprint arXiv:2009.03300},
  year={2020}
}

@inproceedings{pythia,
  title={Pythia: A suite for analyzing large language models across training and scaling},
  author={Biderman, Stella and Schoelkopf, Hailey and Anthony, Quentin Gregory and Bradley, Herbie and O’Brien, Kyle and Hallahan, Eric and Khan, Mohammad Aflah and Purohit, Shivanshu and Prashanth, USVSN Sai and Raff, Edward and others},
  booktitle={International conference on machine learning},
  pages={2397--2430},
  year={2023},
  organization={PMLR}
}

@misc{qwen25,
      title={Qwen2.5 Technical Report}, 
      author={Qwen and : and An Yang and Baosong Yang and Beichen Zhang and Binyuan Hui and Bo Zheng and Bowen Yu and Chengyuan Li and Dayiheng Liu and Fei Huang and Haoran Wei and Huan Lin and Jian Yang and Jianhong Tu and Jianwei Zhang and Jianxin Yang and Jiaxi Yang and Jingren Zhou and Junyang Lin and Kai Dang and Keming Lu and Keqin Bao and Kexin Yang and Le Yu and Mei Li and Mingfeng Xue and Pei Zhang and Qin Zhu and Rui Men and Runji Lin and Tianhao Li and Tianyi Tang and Tingyu Xia and Xingzhang Ren and Xuancheng Ren and Yang Fan and Yang Su and Yichang Zhang and Yu Wan and Yuqiong Liu and Zeyu Cui and Zhenru Zhang and Zihan Qiu},
      year={2025},
      eprint={2412.15115},
      archivePrefix={arXiv},
      primaryClass={cs.CL},
      url={https://arxiv.org/abs/2412.15115}, 
}

@article{mamba,
  title={Mamba: Linear-time sequence modeling with selective state spaces},
  author={Gu, Albert and Dao, Tri},
  journal={arXiv preprint arXiv:2312.00752},
  year={2023}
}


\end{document}